\documentclass[conference]{IEEEtran}
\usepackage{times}

\usepackage[bookmarks=true]{hyperref}
\usepackage[numbers]{natbib}
\usepackage{multicol}
\usepackage{amsmath,amssymb}
\usepackage{algorithmic}
\usepackage{algorithm}
\usepackage{graphicx}
\usepackage[utf8]{inputenc}
\usepackage{pgfplots}
\usepackage{listings}
\usepackage{overpic}

\makeatletter
\let\old@ps@headings\ps@headings
\let\old@ps@IEEEtitlepagestyle\ps@IEEEtitlepagestyle
\def\confheader#1{%
\def\ps@headings{%
\old@ps@headings%
\def\@oddhead{\strut\hfill#1\hfill\strut}%
\def\@evenhead{\strut\hfill#1\hfill\strut}%
}%
\def\ps@IEEEtitlepagestyle{%
\old@ps@IEEEtitlepagestyle%
\def\@oddhead{\strut\hfill#1\hfill\strut}%
\def\@evenhead{\strut\hfill#1\hfill\strut}%
}%
\ps@headings%
}
\makeatother

\confheader{%
To appear in Proceedings of Robotics: Science and Systems,
June 22-26, Freiburg im Breisgau, Germany, 2019}

\begin{document}

\newlength\figureheight
\newlength\figurewidth

\pdfinfo{
  /Author (Michael Burke)
  /Title  (From explanation to synthesis: Compositional program induction for learning from demonstration)
  /CreationDate (D:20191201120000)
  /Subject (Robots)
  /Keywords (Inference, Hybrid systems, Learning from demonstration)
}

\title{From explanation to synthesis: Compositional program induction for learning from demonstration}




%
\author{\authorblockN{Michael Burke\authorrefmark{1},
Svetlin Penkov\authorrefmark{2},
Subramanian Ramamoorthy\authorrefmark{1}\authorrefmark{2}}
\authorblockA{\authorrefmark{1}Institiute for Perception action and Behaviour\\
School of Informatics, University of Edinburgh}
\authorblockA{\authorrefmark{2}FiveAI\\
Email: \{mburke33, sv.penkov, s.ramamoorthy\}\@ed.ac.uk}}

\maketitle

\begin{abstract}
Hybrid systems are a compact and natural mechanism with which to address problems in robotics. This work introduces an approach to learning hybrid systems from demonstrations, with an emphasis on extracting models that are explicitly verifiable and easily interpreted by robot operators. We fit a sequence of controllers using sequential importance sampling under a generative switching proportional controller task model. Here, we parameterise controllers using a proportional gain and a visually verifiable joint angle goal. Inference under this model is challenging, but we address this by introducing an attribution prior extracted from a neural end-to-end visuomotor control model. Given the sequence of controllers comprising a task, we simplify the trace using grammar parsing strategies, taking advantage of the sequence compositionality, before grounding the controllers by training perception networks to predict goals given images. Using this approach, we are successfully able to induce a program for a visuomotor reaching task involving loops and conditionals from a single demonstration and a neural end-to-end model. In addition, we are able to discover the program used for a tower building task. We argue that computer program-like control systems are more interpretable than alternative end-to-end learning approaches, and that hybrid systems inherently allow for better generalisation across task configurations.
\end{abstract}


\section{Introduction}

Recent work in end-to-end learning has resulted in significant advances in the synthesis of visuomotor robot controllers. However, many of these approaches require extensive amounts of data for training, and there are concerns regarding the reliability and verifiability of controllers obtained using these approaches. This is particularly true in industrial settings, where robot operators require interpretable systems with easily verifiable behaviours, and which can be reconfigured with relative ease when minor task changes occur.

As robotics moves beyond task-level learning and starts to address more challenging scenarios, particularly those requiring memory and conditional perception-action loops, end-to-end models are required to become more and more complex, and as a result are less interpretable and more difficult to alter for different scenarios. Addressing these challenges often requires capturing additional data and model retraining, a particularly time consuming exercise.

Hybrid systems \cite{goebel2009hybrid} are a natural remedy to these challenges, and have a long history of application in robotics. This paper seeks to learn hybrid systems from one-shot demonstrations, inferring high-level switching logic that can be easily adjusted by a robot operator, while ensuring that lower-level motion primitives correspond to controllers that are easily verified and inspected, but retaining the benefits of neural learning through perceptual grounding networks.
\begin{figure}
    \centering
    \includegraphics[width=\linewidth]{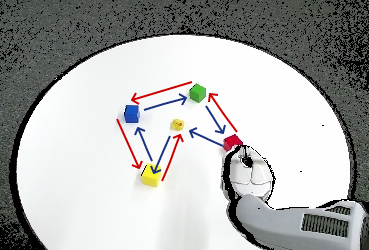}
    \caption{We consider a continual inspection task where a PR2 robot is required to inspect four coloured blocks in a particular order, as indicated by the red arrows above. Upon reaching the rubber duck, the robot is required to reverse its inspection direction, as indicated by the blue arrows.\label{fig:task}}
\end{figure}

In order to achieve this, we propose building a hybrid system in a bottom-up fashion, using a generative switching proportional controller model to identify sub-controllers that constitute a demonstration, and inferring a high-level program that executes these controllers in an appropriate manner to reproduce the observed behaviour. Our motivation for this is to align with sequential composition theories in robotics \cite{Burridge99}, where tasks are solved by moving between sub-controllers lying within the domains of one another. Under our inference framework, each sub-controller is represented by a goal (a joint angle configuration) and an appropriate proportional gain. These sub-controllers can be grounded using a perception network, which predicts goals for a given scene and controller. Importantly, this framework allows for goals to be visually verified by a robot operator, and follows an established perception-action control loop. We use proportional controllers not only for the level of interpretability they provide, but because they allow very rich behaviours to be constructed when chained. Pose-goal-based task representations naturally allow for the use of planning and collision avoidance schemes for safe execution in manipulation tasks. Moreover, grounded proportional control systems can be considered to be visual-servoing instances of the funnelling compositions of \citet{Burridge99}, and are thus a natural mechanism with which to investigate visuomotor control.

Our primary contribution is a method for extracting a programmatic hybrid dynamical system from models trained using end-to-end learning, allowing for the flexibility and interepretability of hybrid control to be used in conjunction with end-to-end learning, and facilitating generalisation across task configurations. In line with this:
\begin{itemize}
    \item We introduce a probabilistic generative task model that infers proportional controller motion primitives from demonstrations and by performing sensitivity analysis on end-to-end models.
    \item We bridge statistical learning and symbolic reasoning in order to move beyond low-level task learning, and allow for programmatic reasoning that facilitates generalisation across task configurations.
    \item We incorporate kinematic structure into the inference process to ensure that inferred symbols map to motion primitives and concepts that are easily verified by robot operators.
\end{itemize}

We demonstrate this method on a task (Figure \ref{fig:task}) requiring the identification of loops, conditionals and low-level motion primitives.

\section{Related Work}

Learning from demonstration (LfD) \cite{ARGALL2009469} is widely acknowledged as a particularly useful paradigm for robot programming. Significant progress has been made in LfD, moving beyond the direct replication of motions to produce more robust approaches \cite{atkeson1997robot} through the introduction of more general schemes for modelling motion like dynamic motion primitives \cite{pastor2009learning}, linear dynamical attractor systems \cite{Dixon04}, sparse online Gaussian processes \cite{Butterfield10,grollman2008sparse} or conditionally linear Gaussian models \cite{Chiappa10,levine2014learning} that can be used for trajectory optimisation.

More recently, trajectory optimisation approaches have been extended to incorporate end-to-end learning, demonstrating robust task level visuomotor control \cite{levine2016end} through guided policy search. End-to-end learning has allowed for the use of domain transfer to facilitate one-shot learning \cite{Yu-RSS-18} from human video demonstrations, and for the use of reinforcement learning to learn optimised control policies \cite{Rajeswaran-RSS-18,Zhu-RSS-18}.

Unfortunately, end-to-end learning approaches typically lack interpretability, are difficult to verify without policy distillation \cite{bastani2018verifiable} and require significant amounts of training data \cite{pinto2016supersizing}. In addition, many of these approaches have been criticised as only seeking to imitate and reproduce motions, with little thought to higher level conceptual learning and reasoning \cite{lake2015human,Lazaro-Gredillaeaav3150}. Work on options learning \cite{sutton1999between, konidaris2009skill} and skill identification \cite{Niekum11,ranchod2015nonparametric} has paid greater attention to this aspect, but has arguably slowed with a recent emphasis on end-to-end learning. 

End-to-end learning systems often lack flexibility, for example if a task goal is modified \cite{lake2017building} or when systematic compositional differences between test and training data are present \cite{marcus2001algebraic}. Hybrid systems combining the high level symbolic reasoning (long upheld as an essential ingredient of cognition) with sub-symbolic machine learning systems are a natural means of addressing these challenges \cite{lake_ullman_tenenbaum_gershman_2017}. For robotics, a key consideration in leveraging symbolic reasoning is the symbol grounding problem \cite{HARNAD1990335}, which seeks to relate high level conceptual reasoning to physical phenomena or behaviours. A number of symbol grounding approaches have been proposed. For example, \citet{Dantam12} tokenise human demonstrations using a grammar based on detected object connections, and manually link these to robot controllers for an assembly task. \citet{Penkov17} use eye-tracking to identify symbol locations and facilitate instance learning and symbol grounding. \citet{Lazaro-Gredillaeaav3150} learn high level concepts (programs) by inducing linear sequences of instructions using example programs, given a known set of atomic instructions for perception and control. 

Unfortunately, much work on concept learning has tended to assume that low level motion primitives are known or specified in advance \cite{penkov2017explaining,Lazaro-Gredillaeaav3150} and neglected aspects related to how best to combine a set of low-level primitives to solve a high-level task. Compositional options learning strategies like skill chaining \cite{konidaris2009skill} or LQR-trees \cite{Tedrake-RSS-09} have sought to address this. This work moves towards addressing the disconnect between options learning, end-to-end learning and symbolic reasoning by extracting compositional low level motion primitives from an end-to-end model and synthesising a program using these.

Our work is similar to \citet{Niekum12}, who apply concept learning using a Beta-Process Autoregressive Hidden Markov Model (BP-AR-HMM) to extract high-level skills from demonstrations and then use dynamic motion primitives to model the low-level motions associated with these skills. They extend this work by learning a finite state automaton for skill sequencing, grounding skills with external visual observations for greater flexibility \cite{Niekum-RSS-13,niekum2015learning}. However, our work differs by extracting an explicit program from demonstrations and in that it pays specific attention to the recovery of both compositional and interpretable motion primitives from neural visuomotor controllers, allowing for its potential application as an explanatory mechanism for models trained using end-to-end learning. 

Programmatic structure is a particularly useful means of capturing human-like concepts \cite{lake2015human} that provides strong inductive bias for learning symbolic representations \cite{Penkov17, penkov2019learning, gaunt2017}. Discovering programmatic structure from observations is a challenging problem that has been studied in the domains of grammar inference \cite{higuera2010grammatical}, program synthesis \cite{gulwani2017program}, programming by demonstration \cite{billard2008robot} and end-user programming \cite{lieberman2006enduser}. Learning and working with programs not only enables model checking and interpretation of black box policies \cite{verma2018programmatically, penkov2017explaining}, but also results in powerful abstractions allowing knowledge transfer to novel environments. 

\section{Synthesising interpretable controllers}

As discussed previously, our goal is to synthesise a program comprising low-level motion primitives that are easily interpretable by a robot operator. We accomplish this using a bottom-up approach, by first inferring low-level controllers for a given demonstration using a probabilistic generative task model. These low level controllers are then grounded using controller specific perception networks, so as to facilitate generalisation across task configurations. Finally, trace level program induction is used to extract a program from the sequence of controller primitives inferred using the generative task model. Each of these components is discussed below.

We test this approach using a reaching task in the spirit of \citet{Lazaro-Gredillaeaav3150}. Here, we require our robot to continually inspect a set of 4 coloured blocks and a toy duck in a known order. However, upon reaching the duck, the robot is expected to reverse its inspection direction, as is illustrated in Figure \ref{fig:task}. Although seemingly simple, this task comprises both loops and conditionals, in addition to low level control primitives, and so is a challenge for many existing approaches to learning from demonstration.

\subsection{Generative task models}

Taking inspiration from the work of \citet{Burridge99} on compositional funnels, we model high-level robot tasks using a switching task model comprising sub-tasks described by a set of proportional controllers acting on joint angles $\boldsymbol{\theta}$. Here, we assume that any robot task can be completed using a sequence $(j=1 \hdots J)$ of proportional controllers determining robot joint velocities,
\begin{equation}
\mathbf{u}_{\boldsymbol{\theta}} = K^j_p (\boldsymbol{\theta} - \boldsymbol{\theta}^j_d),
\end{equation}
parameterised by joint angle goals $\boldsymbol{\theta}^j_d$ and controller gains $K^j_p$. This approach contrasts with those used in \cite{levine2016end} (a conditioned linear Gaussian transition model) or the Beta process auto-regressive hidden Markov model \cite{niekum2015learning} (a compact vector auto-regressive process), in that these low level motion primitives are inherently interpretable, that is, the $j^{th}$ sub-controller can be directly inspected by a robot operator, with goals verified for feasibility.

At any stage in a robot task demonstration, a robot may switch to a new controller, $j+1$, or continue towards a goal defined by the current controller, $j$. We model this switching behaviour using a Bernoulli trial, sampling goals and gains from a prior distribution $\mathbf{\Phi}(\boldsymbol{\theta})$ if a switch has occurred, or from Gaussian jitter if no switch has occurred. This produces the generative task model,
\begin{eqnarray}
k &\sim& \text{Bernoulli}(p)\\
\boldsymbol{\theta}^j_d(t) &\sim& \begin{cases}
    \mathcal{N}(\boldsymbol{\theta}^j_d(t-1),\mathbf{Q}_{\boldsymbol{\theta}}) & \text{if } k = 0\\
    \mathbf{\Phi}(\boldsymbol{\theta}) & \text{if } k = 1
\end{cases}\\
K^j_p(t) &\sim& \mathcal{N}(K^j_p(t-1),\mathbf{Q}_{kp})\\
\mathbf{u}_{\boldsymbol{\theta}} &\sim& \mathcal{N}(K^j_p (\boldsymbol{\theta} - \boldsymbol{\theta}^j_d),\mathbf{R}).
\end{eqnarray}

Here, $\mathbf{Q}_{\boldsymbol{\theta}}$ and $\mathbf{Q}_{kp}$ are transition uncertainty terms, while $\mathbf{R}$ represents the control measurement noise. $p$ denotes the probability of switching to a new controller. Given this model and a task demonstration (a sequence of $T$ joint angles and velocities) we infer controller parameters using sequential importance sampling, as described in Algorithm \ref{alg:sis}.

\begin{algorithm}
\caption{Inferring controllers using Sequential importance sampling re-sampling\label{alg:sis}}
\begin{algorithmic}
\STATE Initialise $N$ particles
\FOR{$t=0$ \TO $T$}
\FOR{$k=0$ \TO $pN$}
\STATE Sample $\boldsymbol{\theta}^k_d(t) \sim  \mathcal{N}(\boldsymbol{\theta}^k_d(t-1),\mathbf{Q}_{\boldsymbol{\theta}})$
\STATE Sample $K^k_p(t) \sim \mathcal{N}(K^k_p(t-1),\mathbf{Q}_{kp})$
\STATE Evaluate $L^k = \mathcal{N}(K^j_p (\boldsymbol{\theta} - \boldsymbol{\theta}^j_d),\mathbf{R})$
\ENDFOR
\FOR{$k=pN$ \TO $N$}
\STATE Sample $\boldsymbol{\theta}^k_d(t) \sim \mathbf{\Phi}(\boldsymbol{\theta})$
\STATE Sample $K^k_p(t) \sim \mathcal{N}(K^k_p(t-1),\mathbf{Q}_{kp})$
\STATE Evaluate $L^k = \mathcal{N}(K^j_p (\boldsymbol{\theta} - \boldsymbol{\theta}^j_d),\mathbf{R})$
\ENDFOR
\STATE Draw $N$ samples: $\boldsymbol{\theta}^k_d,K^k_p \sim L^k$
\ENDFOR
\end{algorithmic}
\end{algorithm}

The generative task model described above places no limits on the number of controllers required by a task, and is thus flexible enough to extend to almost any feasible manipulation task, through the sequential composition of controllers. 

\subsection{Attribution priors}

Inference under the model above can be challenging, as there are numerous joint angle goal possibilities to consider, particularly for robots with large work-spaces. As a result, special attention needs to be paid to the switching prior, $\mathbf{\Phi}(\boldsymbol{\theta})$.

Here, we propose a switching prior based on sensitivity information extracted from an end-to-end model trained to predict robot controls given robot states and images. As mentioned previously, end-to-end learning has been shown to be effective for many robot control tasks, but is often undesirable for robot operators and in industrial applications, given the lack of interpretability of these systems. This lack of interpretability extends across much of modern machine learning, and has led to the development of sensitivity analysis techniques that seek to identify which inputs contribute the most to a model's output \cite{selvaraju2017grad,zeiler2014visualizing}. Gradient-based sensitivity maps are a particularly powerful means of extracting the image content that is salient to a deep learning model, and reliably highlight image content of interest for well trained models, even in cluttered scenes.

\begin{figure}
    \centering
    \includegraphics[width=\linewidth]{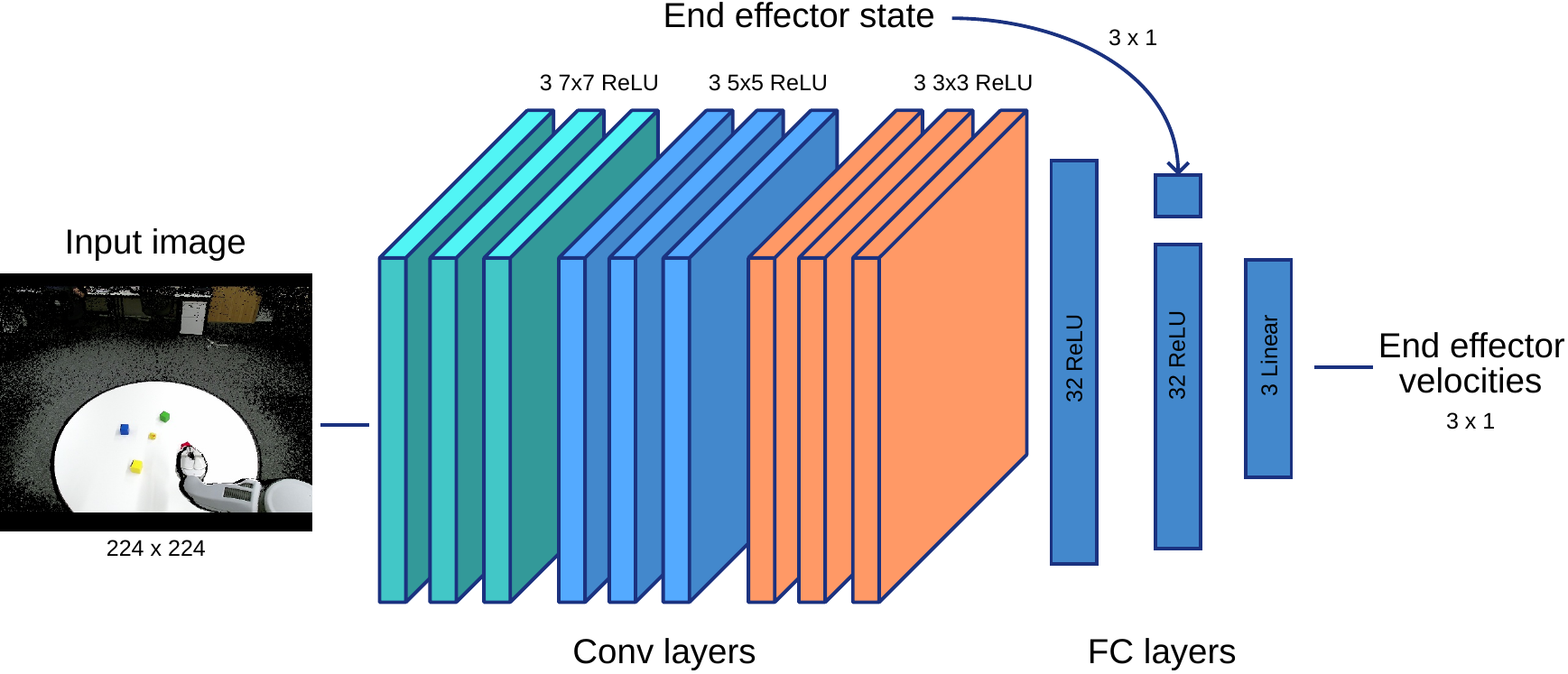}
    \caption{A simple end-to-end model trained (Adam optimiser, learning rate of 1e-2) to predict end effector given images and joint states is used to provide a saliency prior of positions of interest in a scene. This prior is then used to sample potential controller goals.\label{fig:end2end}}
\end{figure}

\begin{figure}
    \centering
    \includegraphics[width=\linewidth]{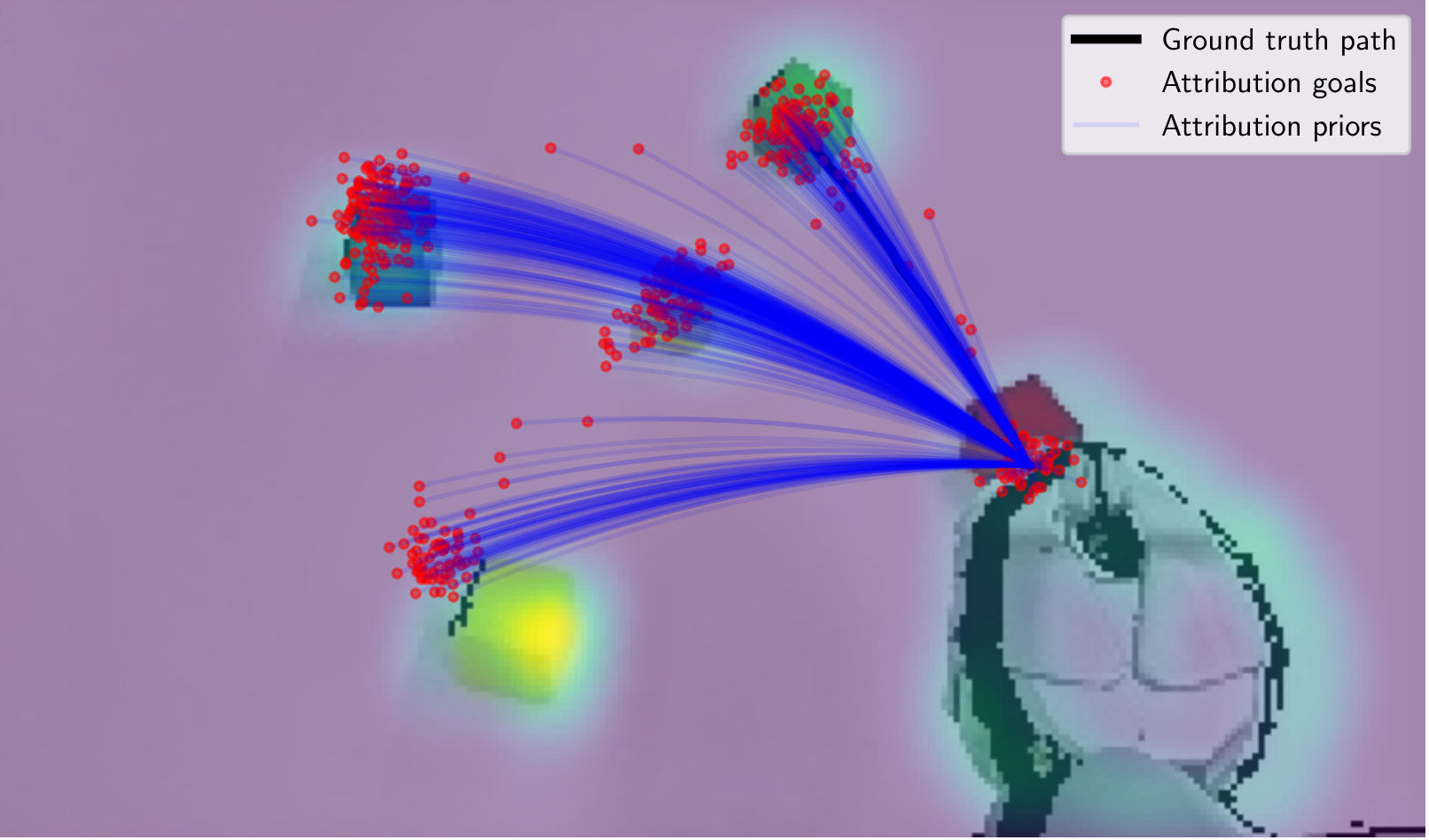}
    \caption{Controller roll-outs to joint states drawn from attribution priors reduce the space of possible joint angle goals. Goal projections are marked in red, with roll-outs in blue. The saliency map used for sampling is extracted by performing sensitivity analysis on the end-to-end prediction model depicted in Figure \ref{fig:end2end}. \label{fig:saliency}}
\end{figure}

We make use of these techniques to construct a prior over controller goals. First, we train an end-to-end prediction model $g(\mathbf{I},\boldsymbol{\theta})$ using 1700 velocity-image pairs, following the architecture in Figure \ref{fig:end2end}. This model is unable to solve our conditional sequential reaching task as it lacks memory, but is still able to identify image regions of potential importance to a prediction, as shown in the saliency map of Figure \ref{fig:saliency}. Our hypothesis is that many of these salient regions correspond to joint goals of importance and that these saliency maps provide a useful prior over regions of interest in a scene.

With this in mind, we sample joint angles using a saliency map and the rejection sampling described in Algorithm \ref{alg:rej_sample}. Here, goal samples are intially drawn from the set of $M$ joint angles in the demonstration sequence, starting from current joint state $\boldsymbol{\theta}(t)$. This restricts joint angle goals to future joint angles in the demonstration. These samples are then projected into the image plane using robot forward kinematics and a camera model, and an attribution likelihood 
\begin{equation}
L_a^k = \frac{\partial g(\mathbf{I},\boldsymbol{\theta}(t))}{\partial \mathbf{I}(\mathbf{x}^k_{e})}
\end{equation} 
is evaluated  at this image coordinate, $\mathbf{x}^k_{e}$. Finally, we re-sample and jitter the sampled angles given this likelihood to obtain joint angle goal samples.
\begin{algorithm}
\caption{Sampling from attribution priors\label{alg:rej_sample}}
\begin{algorithmic}
\STATE Draw $k = 1 \hdots N_p$ samples: $\boldsymbol{\theta}^k_d \sim \left [ \boldsymbol{\theta}(t) \hdots \boldsymbol{\theta}(t+M)\right]$\\

\STATE Project samples into image plane: $\mathbf{x}^k_{e} = \mathbf{K} \text{ FK}(\boldsymbol{\boldsymbol{\theta}}^k_d)$\\
\STATE Evaluate attribution likelihood: $L_a^k = \frac{\partial g(\mathbf{I},\boldsymbol{\theta}(t))}{\partial \mathbf{I}(\mathbf{x}^k_{e})}$\\
\STATE Draw $k = 1 \hdots N_p$ samples: $\boldsymbol{\theta}^k_d \sim L_a^k$
\end{algorithmic}
\end{algorithm}

Figure \ref{fig:saliency} shows projected joint angle goal samples drawn using the sampling process above, along with controller roll-outs towards these. It is clear that the proposed sampling process is able draw upon the visual information captured by the end-to-end prediction model to provide good priors over joint angle goals and potential destinations. It should be noted that the saliency prior and proportional goal configuration inductively bias the inference towards goals that correspond with specific objects present in the scene. As a result, saliency-based sampling is unlikely to be useful in situations where image content causes repulsion, for example in a collision avoidance setting, and may result in multiple unused particles here.

\subsection{Symbol extraction}

\begin{figure}
    \centering
    \setlength\figureheight{4.5cm}
    \setlength\figurewidth{5.5cm}
\begin{tikzpicture}

\begin{axis}[
colorbar,
colorbar style={ytick={-6,-4,-2,0,2},yticklabels={-6.0,-4.0,-2.0,0.0,2.0},ylabel={}},
colormap/viridis,
height=\figureheight,
point meta max=3.36337411592398,
point meta min=-7.67805151251406,
tick align=outside,
tick pos=left,
width=\figurewidth,
x grid style={lightgray!92.02614379084967!black},
xlabel={Samples},
xmin=-0.5, xmax=4482.5,
y grid style={lightgray!92.02614379084967!black},
ymin=-0.5, ymax=8.5,
ytick={0,1,2,3,4,5,6,7,8},
yticklabels={Torso lift  goal,Shoulder pan  goal,Shoulder lift  goal, Upper arm roll  goal, Elbow flex  goal, Forearm roll  goal, Wrist flex  goal, Wrist roll  goal, Controller gain}
]
\addplot graphics [includegraphics cmd=\pgfimage,xmin=-0.5, xmax=4482.5, ymin=8.5, ymax=-0.5] {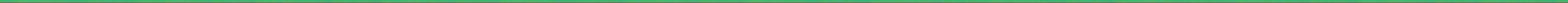};
\end{axis}

\end{tikzpicture}
    \caption{A distribution over joint angle goals and controller gains is obtained at each time step in the sequence. The figure shows the maximum-a-posteriori estimates for each controller parameter (y-axis) at each time step. The colour bar indicates the inferred parameter values. (Figure best viewed electronically).\label{fig:seq}}
\end{figure}

Inference using sequential importance sampling and the attribution prior produces a distribution over possible controllers for each time step in a demonstration sequence (Figure \ref{fig:seq}). Importantly, the periodicity in inferred controller parameters is clearly visible. However, our goal is to build a high level sequence of controllers that describe the demonstrated task at a symbolic level. We isolate this sequence of controllers constituting a demonstration by making use of the effective particle size \cite{kong1994sequential}, 
\begin{equation}
N_{eff} = \frac{1}{\sum_{k=0}^N{(L^k)^2}},
\end{equation} 
at each time step of the inference process. The intuition behind this is as follows: When a demonstration requires a controller switch towards a new goal, a significant portion of the particles will have low probability mass and therefore a low effective particle size. In contrast, the effective particle size will be greatest when only a single controller is required and a clear joint goal and gain has been established. 

\begin{figure}
    \centering
    \setlength\figureheight{5cm}
    \setlength\figurewidth{8.8cm}
    \input{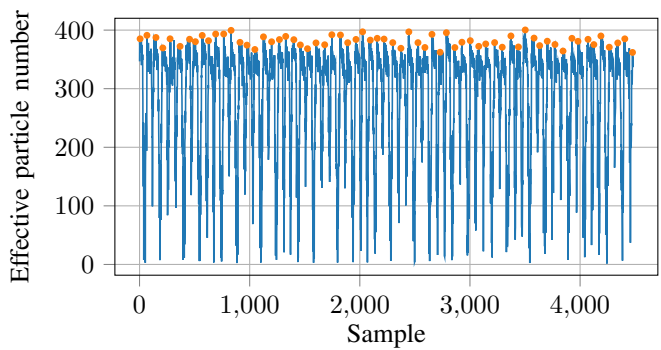}
    \caption{The effective particle size is used to identify the chain of proportional controllers used in a demonstration sequence. \label{fig:Neff}}
\end{figure}

This is illustrated for our running example in Figure \ref{fig:Neff}, which shows the effective particle size for each sample in the demonstration sequence. A simple peak detector is able to identify points at which a clear joint goal and gain are present, which typically occurs mid-way through a motion. We construct a sequence of controllers by selecting the maximum-a-posteriori controller at each of these points in the demonstration.

\begin{figure}
    \centering
    \setlength\figureheight{2cm}
    \setlength\figurewidth{10cm}
\begin{tikzpicture}

\begin{axis}[
height=\figureheight,
tick align=outside,
tick pos=left,
width=\figurewidth,
x grid style={lightgray!92.02614379084967!black},
xlabel={Controller sequence},
xmin=-0.5, xmax=64.5,
y grid style={lightgray!92.02614379084967!black},
ymin=-0.5, ymax=0.5,
ymajorticks=false
]
\addplot graphics [includegraphics cmd=\pgfimage,xmin=-0.5, xmax=64.5, ymin=0.5, ymax=-0.5] {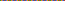};
\end{axis}

\end{tikzpicture}
    \includegraphics[width=\linewidth]{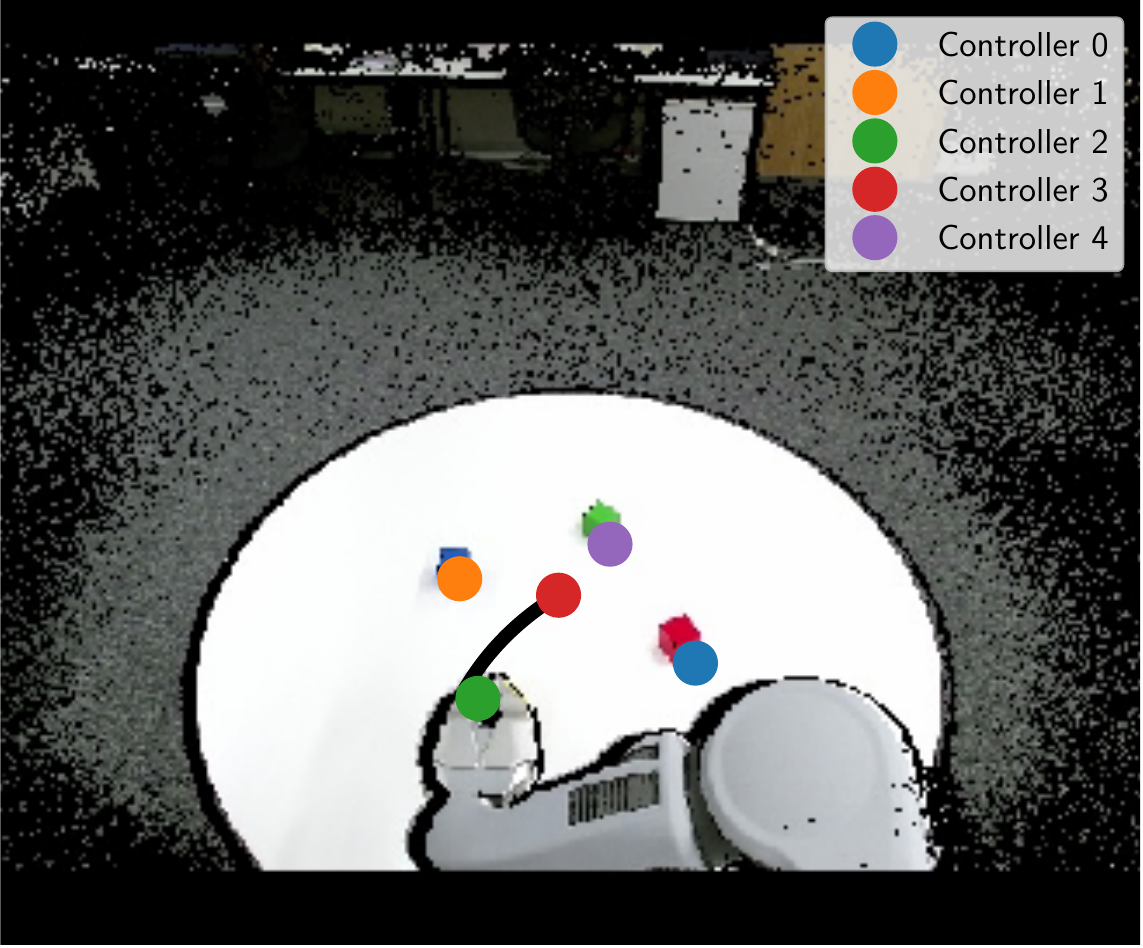}
    \caption{Maximum-a-posteriori controllers are clustered to produce a symbolic trace of controller primitives comprising a demonstration. The black line in the figure illustrates a roll out from the current robot state using Controller 3.\label{fig:goals}}
\end{figure}

This sequence of controller primitives is then clustered ($K$-means clustering using an elbow criterion to select the number of clusters) in order to produce a symbolic behaviour trace for the demonstration, as shown at the top of Figure \ref{fig:goals}. Here, each controller is represented by a distinct colour and the sequence of goals constituting a task is clearly visible.

Visual inspection of the image projection of the joint angle goals associated with the identified controller primitives clearly shows that we have successfully extracted the key primitives making up our demonstration sequence, along with the order in which these were visited. Importantly, the task model structure used for inference means that these goals are verifiable and easily interpretable by a robot programmer. 

\subsubsection{Comparison with linear Gaussian systems}

We contrast this with the linear motion primitives learned using the conditional Gaussian mixture model approach \cite{levine2016end}, which lack a clear interpretation. These take the form 
\begin{equation}
\boldsymbol{\theta}(t) = \mathbf{A}^j \boldsymbol{\theta} (t-1) + \mathbf{B}^j \mathbf{u}_{\boldsymbol{\theta}},
\end{equation}
and require that a linear quadratic regulator be used to find controls given the current state. 

This formulation is not particularly compact, as it requires multiple models to describe the motion towards a particular goal. For example, when this approach was used to model motions in our running example, \textbf{15} models were required to describe trajectories, using the Bayesian information criterion for model selection. In contrast, the proposed task model produces \textbf{5} clearly interpretable models. Fewer models are required for the latter because the proportional controller formulation is invariant to the direction from which a goal is approached. 

\subsubsection{Evaluating the effects of attribution}
\begin{figure}
    \centering
    \setlength\figureheight{5cm}
    \setlength\figurewidth{8cm}
    \input{./figs/Neff_comparison.tikz}
    \includegraphics[width=0.8\linewidth]{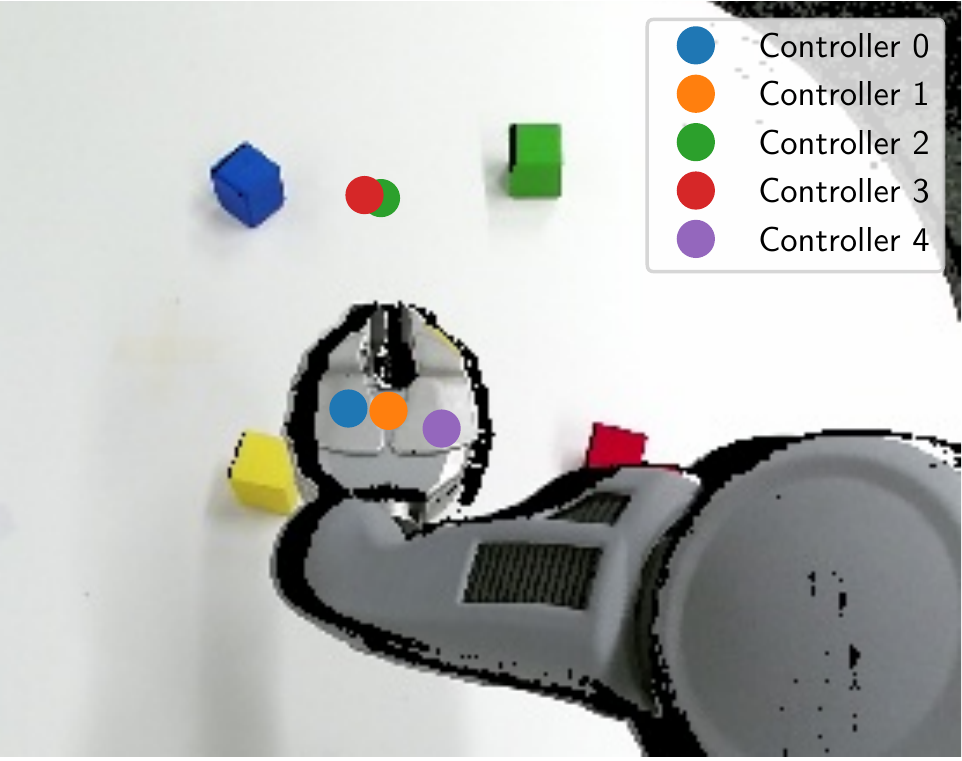}
    \caption{The effective particle size is significantly improved by the inclusion of attribution information, indicating improved inference. Inference without this information fails to extract sensible controller goals. \label{fig:Neff_comparison}}
\end{figure}

We compared inference using visual attribution priors to a particle filter under the proposed generative model, but with the sensitivity-based rejection sampling step replaced by a direct draw from the $M$ future demonstration states directly. As shown in Figure \ref{fig:Neff_comparison}, inference using this approach failed to correctly identify suitable proportional controller goals. This can be attributed to a poor effective sample size (see top of Figure \ref{fig:Neff_comparison} and Table \ref{tab:my_label}), which makes it challenging to identify controller switching and infer controllers. 

\begin{table}[h]
    \centering
    \caption{Inference quality \label{tab:my_label}}
    \begin{tabular}{|c|c|c|c|c|}
        \hline
          \textbf{Statistics: $N_{eff}$} & \textbf{Mean} & \textbf{Max} & \textbf{Min} & \textbf{IQR}\\
         \hline
         Attribution ($N_p = 50$) & 32.12 & 48.73 & 3.80 & 12.45 \\
         \hline
         Baseline ($N_p = 50$) & 28.12 & 42.96 & 1.00 & 7.60 \\
         \hline
    \end{tabular}
\end{table}

Importantly, the improved inference quality introduced by the attribution prior results in a greater interquartile range in effective particle number, which allows for the improved detection of controller switches. These experiments clearly show that the proposed rejection sampling approach extracts visual information from the end-to-end model to inform the task identification process.

\subsection{Program induction}

The symbolic trace identified above can be simplified into a programmatic representation that can easily be inspected and modified by a robot operator, thereby providing greater levels of flexibility. As discussed earlier, a number of program induction techniques have been developed previously, to deal with a variety of grammars or program primitives. These approaches can be expensive and often introduce errors if they fail to reproduce an instruction trace perfectly. In addition, program induction is complicated by the fact that any number of programs could produce a given instruction trace, and the choice of whether one program is more interpretable or elegant than another is highly subjective. The latter is a particular challenge for program induction, as the solution space for programs is extremely large.

\begin{figure}
    \centering
    \includegraphics[width=0.8\linewidth]{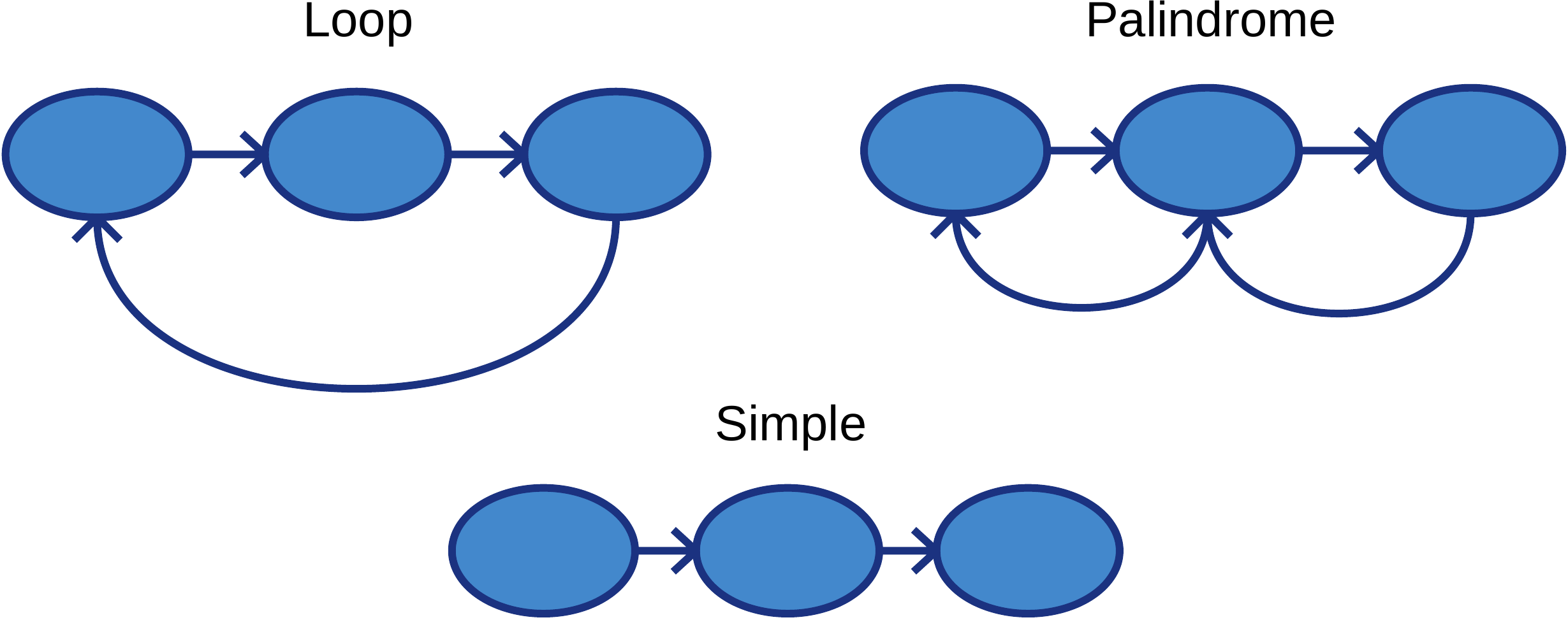}
    \caption{Three key sub-structures are identified to extract a program from symbolic traces. Loops are a sequence of controllers executed cyclically, palindromes are a sequence of controllers executed in a path reversal process, and simple controller sequences have a clear start and end goal.\label{fig:pi_structures}}
\end{figure}

In light of this, we propose a string parsing approach that deals with three key aspects (Figure \ref{fig:pi_structures}) that we consider likely to arise from the compositional structure of the trace inferred using the proposed generative task model. Importantly, these correspond to commonly observed mission requirements in robotics \cite{menghi2019specification}, specifically those required for patrolling and sequential visits. This trace is formed of a sequence of proportional controllers, driving a robot through a series of goal states to solve some task. These motion primitives mean that we are likely to observe the following aspects in general tasks. 

The first is a looping structure, where a cycle of goal states is visited a number of times. This structure can be identified by finding repeated consecutive sub-sequences in a trace. Since our goal is to produce shorter program structures, we condense traces by replacing repeated sub-sequences in a trace with loops, using an exhaustive search to identify repeated sub-sequences, and replacing those sub-sequences that repeat the most with looped instructions. We repeat this process for code segments both outside and inside the replaced loops, until no repeated sub-sequences remain.

The second structure we consider is a palindromic or retreating sequence. Here, a task may require a robot to pass through a set of goal states, and then return by reversing through the same set of goal states. We identify this structure by exhaustively searching through code segments after loop detection, and finding all palindromes in the instruction trace of a code segment. We replace palindromic sub-sequences (only those longer than 6 instructions to avoid unnecessary complication) in code by a controller list and a counter that increments until the palindrome centre, before decrementing until the palindrome ends. The palindrome centre state can be considered a goal condition that needs to be met before a trace continues. 

\definecolor{codegreen}{rgb}{0,0.6,0}
\definecolor{codegray}{rgb}{0.5,0.5,0.5}
\definecolor{codepurple}{rgb}{0.58,0,0.82}
\definecolor{backcolour}{rgb}{0.95,0.95,0.92}

\lstdefinestyle{mystyle}{backgroundcolor=\color{backcolour},
    commentstyle=\color{codegreen},
    keywordstyle=\color{magenta},
    numberstyle=\tiny\color{codegray},
    stringstyle=\color{codepurple},
    basicstyle=\footnotesize,
    breakatwhitespace=false,         
    breaklines=true,                 
    captionpos=t,                    
    keepspaces=true,                 
    numbers=none,                    
    numbersep=5pt,                  
    showspaces=false,                
    showstringspaces=false,
    showtabs=false,                  
    tabsize=2} 
\lstset{style=mystyle}
\begin{lstlisting}[language=Python, caption=Simplified program extracted from trace.,label={lst:program}]

# Looping sequence
c_list = [2,1,4,0,3]
for j in range(6):
    
    # Palindromic sequence
    count = 0
    for k in range(len(c_list)*2-1):
        execute(c_list[count])
        if (k >= len(c_list)-1):
            count = count-1
        else:
            count = count+1
            
    # Simple sequence
    execute(3)
    
# Simple sequence
execute(2)
execute(1)
execute(4)
execute(0)
execute(3)

\end{lstlisting}

Once these structures have been removed, the only remaining code blocks in the program will contain simple sequences with clear starting and terminating goal states. Program listing \ref{lst:program} illustrates this for our running example. It is clear that the synthesised program captures the demonstrated behaviour compactly, and could easily be modified by a robot operator, say to extend the number of loop repetitions or to insert a new goal if required.

This program is both conceptually similar and of equivalent complexity to the hand-coded example originally used to generate the demonstration (see Program listing \ref{lst:program_orig}), which looped indefinitely, executing a simple transition model on each iteration. Differences arise from the finite demonstration length used to infer symbolic traces and the constraints of the proposed triple structure program grammar, which is restricted to loops, palindromes and simple sequences in order to ensure generalisability across tasks.

\begin{lstlisting}[language=Python, caption=Original program used to generate demonstration.,label={lst:program_orig}]

state_explanation = [3,2,1,4,0,3]

class sm():
    def __init__(self):
        self.step = 1
        self.state = 0
    
    def transition(self,state):   
        self.state = self.state+self.step
        if (self.state == 5):
            self.step = -1
        if (self.state == 0):
            self.step = 1
        return self.state
    
seq = sm()
state = 0
while (1):
    state = seq.transition(state)
    mc.move(state_explanation[state])

\end{lstlisting}

\subsection{Symbol grounding}

The previous section showed how we can infer low-level controller primitives and construct a corresponding program describing a robot demonstration using these. However, symbol grounding is required if the inferred symbols are to generalise to different configurations of the demonstrated task.

\begin{figure}
    \centering
    \includegraphics[width=0.8\linewidth]{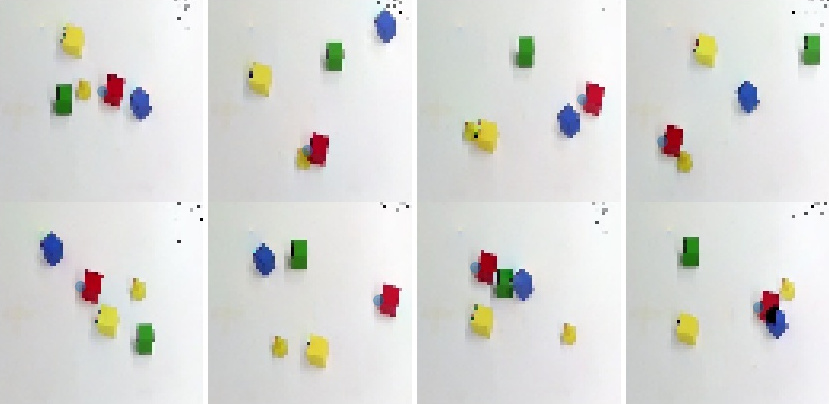}
    \caption{Grounding networks are trained to predict controller goals (joint angle states) using 3200 synthetic images generated by cropping patches around identified controller goal states and pasting these in varying configurations corresponding to projected joint angle states.\label{grounding_data}}
\end{figure}

For the purpose of illustration, we ground our symbols by means of symbol-conditioned perception networks (4 convolutional and two dense fully connected layers), $g_c(\mathbf{I},\boldsymbol{\phi})$, trained to predict controller goal locations for a given scene, using a mean square error loss,
\begin{equation}
J = \mathbb{E}\left[\Vert g_c(\mathbf{I},\boldsymbol{\phi})-\mathbf{y}_{\boldsymbol{\theta}}(c)\Vert\right].
\end{equation}
Here $\mathbf{y}_{\boldsymbol{\theta}}(c)$ denotes the goal angles extracted from the demonstration sequence for controller $c$, and $\boldsymbol{\phi}$ represents network parameters. We leverage the fact that controller goals correspond to visual information for data augmentation, cropping and re-positioning image patches around joint angle goals at projected image locations of joint angles sampled from the robot workspace (Figure \ref{grounding_data}). This augmentation scheme is inductively biased towards object-based goal prediction, implicitly making the assumption that goals are associated with content in image patches around these. Alternative grounding networks may be required for other situations and for better generalisation to scene variation or multi-camera settings, but this is beyond the scope of this work. Goal prediction via more advanced deep neural networks trained using supervised learning has recently shown excellent performance in real-time drone racing \cite{kaufmann18}, and this performance is likely to extend to more complex tasks.

\subsection{Program synthesis}
\begin{figure}
    \centering
    \includegraphics[width=\linewidth]{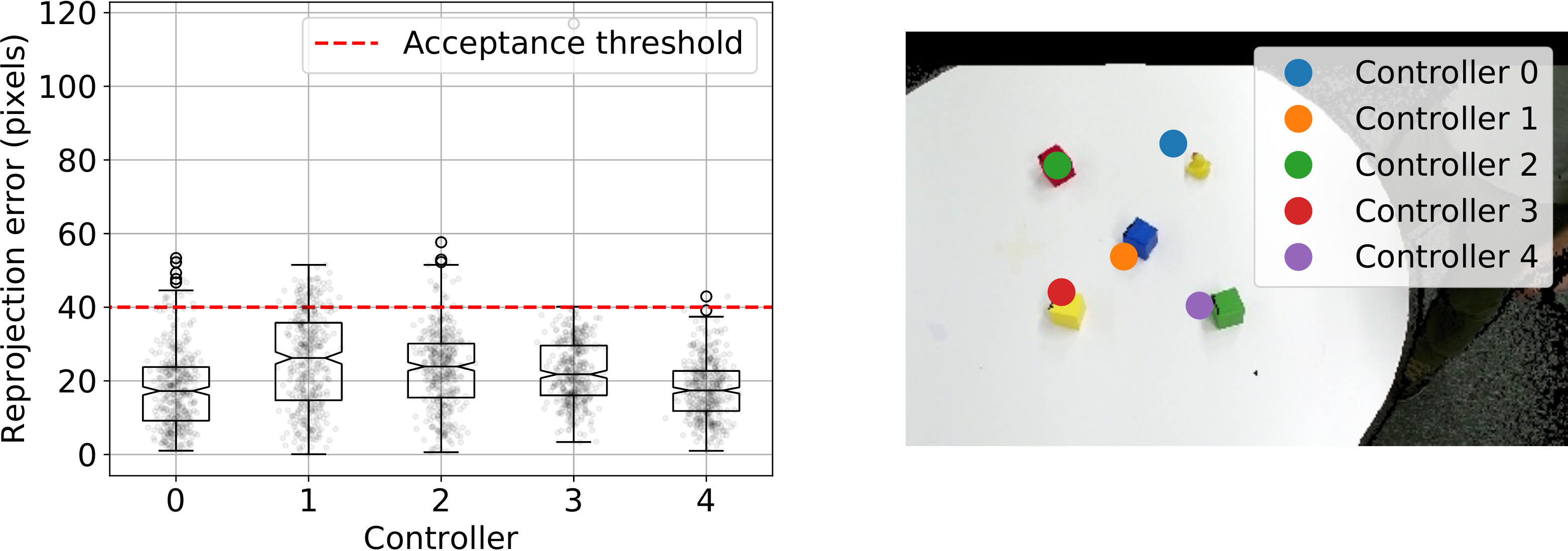}
    \caption{Symbol grounding allows the synthesis of the high level program in different contexts, thereby exhibiting high levels of generalisation. Boxplots of the error between projected joint angle goals and manually labelled positions are shown above for 400 arbitrarily arranged test images. This allows generalisation to previously unseen object configurations. \label{fig:generalise}}
\end{figure}

Given the inferred program and controller primitives, program synthesis becomes trivial. Here, we simply follow the high level program, and execute the required controllers as requested. Controller execution is a simple matter of selecting a joint angle goal using the symbol grounding network, and using a proportional control law with the inferred gains.
\begin{figure*}[!ht]
    \centering
    \begin{overpic}[height=0.175\linewidth,width=0.13\linewidth]{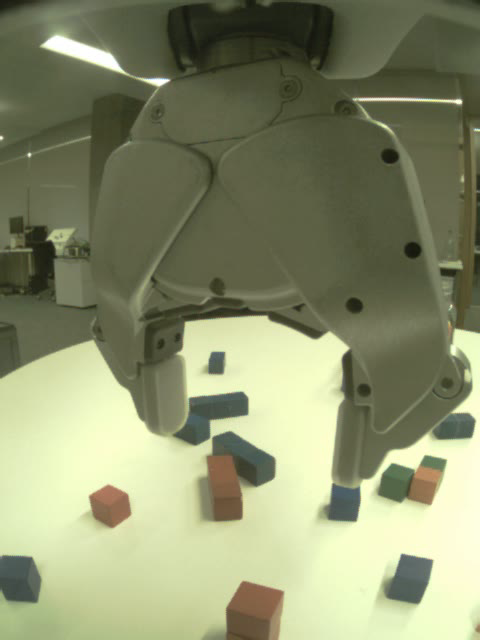}\put(-0.1,85){\colorbox{white}{\strut Above tower 1\hspace{7mm}}}\end{overpic}
    \begin{overpic}[height=0.175\linewidth,width=0.13\linewidth]{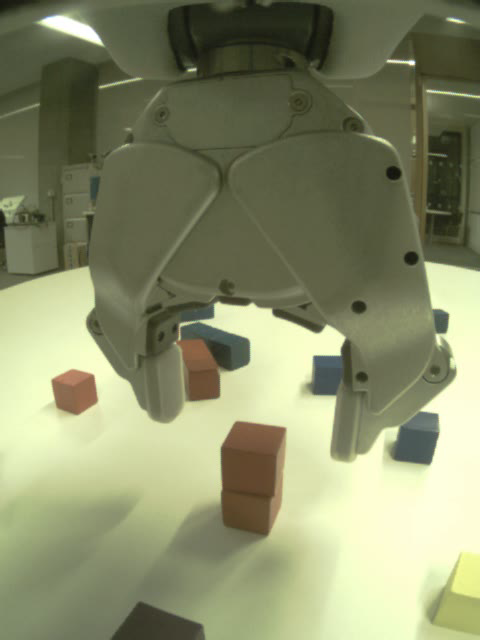}\put(-0.1,85){\colorbox{white}{\strut Lower gripper\hspace{4mm}}}\end{overpic}
    \begin{overpic}[height=0.175\linewidth,width=0.13\linewidth]{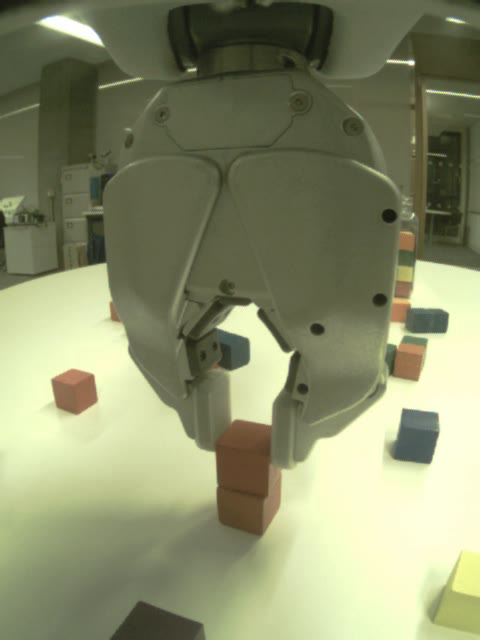}\put(-0.1,85){\colorbox{white}{\strut Close gripper\hspace{4mm}}}\end{overpic}
    \begin{overpic}[height=0.175\linewidth,width=0.13\linewidth]{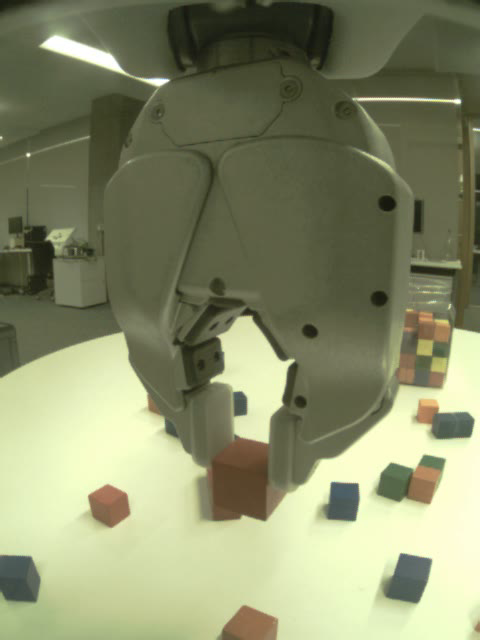}\put(-0.1,85){\colorbox{white}{\strut Lift gripper\hspace{7mm}}}\end{overpic}
    \begin{overpic}[height=0.175\linewidth,width=0.13\linewidth]{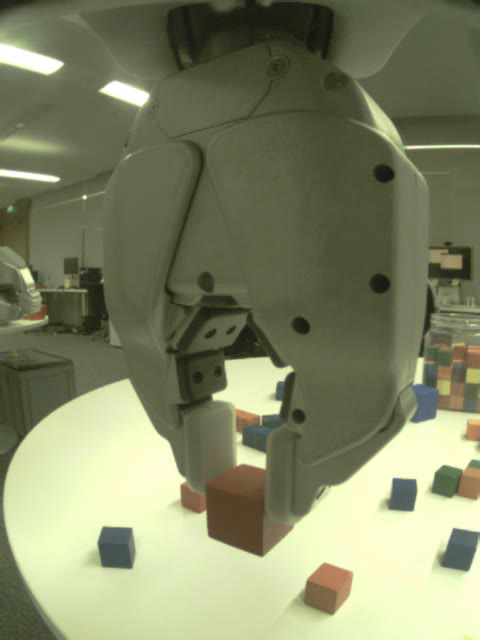}\put(-0.1,85){\colorbox{white}{\strut Above tower 2\hspace{7mm}}}\end{overpic}
    \begin{overpic}[height=0.175\linewidth,width=0.13\linewidth]{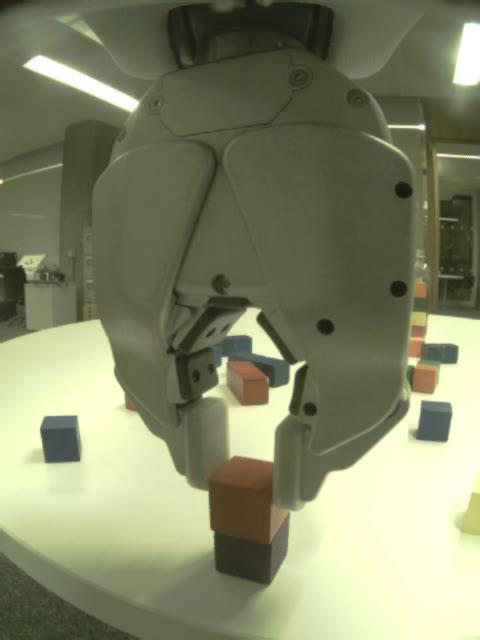}\put(-0.1,85){\colorbox{white}{\strut Lower gripper\hspace{7mm}}}\end{overpic}
    \begin{overpic}[height=0.175\linewidth,width=0.13\linewidth]{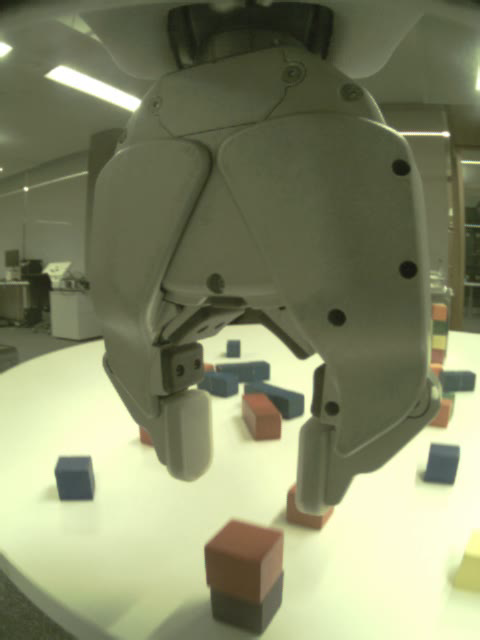}\put(-0.1,85){\colorbox{white}{\strut Open and raise\hspace{7mm}}}\end{overpic}
    \caption{A tower assembly task using the motion primitives pictured in the figure is used to investigate the proposed task decomposition framework. \label{fig:tower1}}
\end{figure*}
\begin{figure}[!ht]
    \centering
    \includegraphics[width=0.8\linewidth]{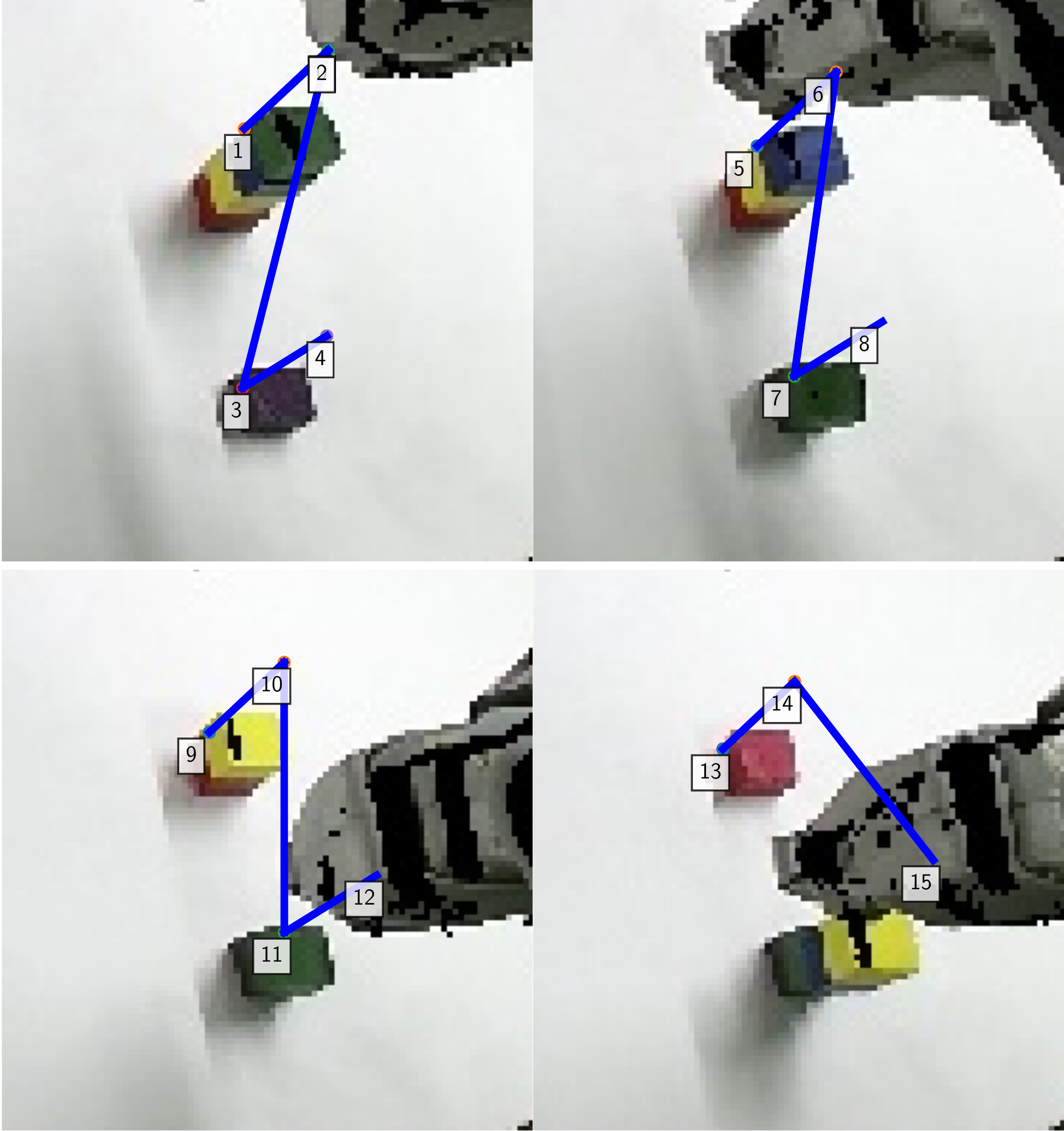}
    \caption{The tower assembly task is decomposed into 4 key controllers (numbered in the figure) using the generative task model. Interestingly, this decomposition is simpler than our original hand-coded rules. \label{fig:tower}}
\end{figure}

This approach allows for generalisation to different task configurations, for example the case in Figure \ref{fig:generalise}, where we have re-ordered the blocks to be visited and the robot is still able to execute the underlying program in the original order. Modifying the program, say to visit blocks in a different order, is easily accomplished using the hybrid system representation. This level of generalisation cannot be easily obtained with end-to-end models like those used for guided policy search \cite{levine2016end}, and the original model (Figure \ref{fig:end2end}) fails in this configuration. Although approaches like this are often able to generalise well to variations within a task, the abstraction provided by learning a grounded program inherently introduces flexibility with regard to alternative task configurations.

It should be noted that the guided policy search model is unable to replicate the task in our running example as it lacks a recurrent memory component, and is thus unable to model repeated palindromic motions, which require motion conditioned on a longer history of states. 

\section{Tower building example}

Thus far, the proposed framework has been demonstrated on a relatively simple reaching task. This section illustrates our approach in a more complex setting, where we inferred a program and controller primitives to explain a tower disassembly and reassembly task. Here, we programmed a PR2 robot to move blocks from one tower to another, relying on hard-coded positions and offsets, following the general rubric indicated in Figure \ref{fig:tower1}: move above block, lower gripper, close gripper, lift gripper, move above goal, lower gripper, open and raise gripper. 

We recorded a single demonstration sequence, and used the generative task model to identify sub-controllers with an end-to-end model trained using supervised learning. Figure \ref{fig:tower} shows the segmented controller goal projections for each sub-task identified. Interestingly, the inferred controllers combine aspects that we had decoupled in code, decomposing each stage of the process above into four components: position gripper, close and raise gripper, position block, open and raise gripper. 

Clustering the identified controllers results in 15 unique controllers, connected in a simple sequence. A key observation is that the proposed programmatic abstraction means that similar concepts, for example placing a block on a tower, are considered different tasks, as they occur in different positions. This highlights the importance of choosing motion primitives that map to likely concepts in a given scenario. Compositional goal-like motion primitives are inductively biased towards the identification of concepts related to inspection or the manipulation of objects, but program abstraction techniques \cite{gulwani2017program} could be applied to identify higher order concepts if necessary.


\section{Conclusion} 
\label{sec:conclusion}

This paper has introduced a method for extracting a program from demonstrations of robot manipulation tasks. We proposed a probabilistic generative model for learning from one-shot demonstrations that decomposes tasks into a sequence of proportional controllers in joint angle space, with corresponding joint angle goals. Inference under this model draws on prior information extracted from an end-to-end visuomotor controller, leveraging sensitivity analysis for goal identification. Program induction was used to reduce the inferred trace into a form that could be easily modified by a robot operator, and perceptual symbol grounding networks trained to allow task generalisation. We argued that this programmatic compositional formulation is inherently interpretable, as joint goals could be verified by inspection, and that the incorporation of structure (kinematic models, knowledge of controller primitives) into the inference process both simplifies and ensures interpretability of learned behaviours. 

Importantly, the proposed approach can be used as an explanation mechanism for reinforcement learning agents or existing end-to-end systems. Synthesising a program by observing an agent can provide valuable insights into its learned behaviours. This also allows for the transfer of learned behaviours when tasks or goals are modified, thereby introducing greater flexibility and improving generalisation across task configurations. Finally, the proposed approach eases system verification, which currently poses a significant barrier to the deployment of end-to-end learning systems.


\section*{Acknowledgements}
This work was supported by funding from the ORCA Hub~EPSRC (EP/R026173/1, 2017-2021) and an Alan Turing Institute funded project, Safe AI for Surgical Assistance.

\bibliographystyle{plainnat}
\bibliography{references}

\begin{thebibliography}{45}
\providecommand{\natexlab}[1]{#1}
\providecommand{\url}[1]{\texttt{#1}}
\expandafter\ifx\csname urlstyle\endcsname\relax
  \providecommand{\doi}[1]{doi: #1}\else
  \providecommand{\doi}{doi: \begingroup \urlstyle{rm}\Url}\fi

\bibitem[Argall et~al.(2009)Argall, Chernova, Veloso, and
  Browning]{ARGALL2009469}
Brenna~D. Argall, Sonia Chernova, Manuela Veloso, and Brett Browning.
\newblock \href{https://doi.org/10.1016/j.robot.2008.10.024}{A survey of robot
  learning from demonstration}.
\newblock \emph{Robotics and Autonomous Systems}, 57\penalty0 (5):\penalty0 469
  -- 483, 2009.
\newblock ISSN 0921-8890.

\bibitem[Atkeson and Schaal(1997)]{atkeson1997robot}
Christopher~G Atkeson and Stefan Schaal.
\newblock
  \href{http://citeseerx.ist.psu.edu/viewdoc/download?doi=10.1.1.54.8531&rep=rep1&type=pdf}{Robot
  learning from demonstration}.
\newblock In \emph{ICML}, volume~97, pages 12--20. Citeseer, 1997.

\bibitem[Bastani et~al.(2018)Bastani, Pu, and
  Solar-Lezama]{bastani2018verifiable}
Osbert Bastani, Yewen Pu, and Armando Solar-Lezama.
\newblock \href{https://arxiv.org/pdf/1805.08328.pdf}{Verifiable Reinforcement
  Learning via Policy Extraction}.
\newblock \emph{arXiv preprint arXiv:1805.08328}, 2018.

\bibitem[Billard et~al.(2008)Billard, Calinon, Dillmann, and
  Schaal]{billard2008robot}
Aude Billard, Sylvain Calinon, Ruediger Dillmann, and Stefan Schaal.
\newblock Robot programming by demonstration.
\newblock In \emph{Springer handbook of robotics}, pages 1371--1394. Springer,
  2008.

\bibitem[Burridge et~al.(1999)Burridge, Rizzi, and Koditschek]{Burridge99}
R.~R. Burridge, A.~A. Rizzi, and D.~E. Koditschek.
\newblock \href{https://doi.org/10.1177/02783649922066385}{Sequential
  Composition of Dynamically Dexterous Robot Behaviors}.
\newblock \emph{The International Journal of Robotics Research}, 18\penalty0
  (6):\penalty0 534--555, 1999.

\bibitem[Butterfield et~al.(2010)Butterfield, Osentoski, Jay, and
  Jenkins]{Butterfield10}
J.~Butterfield, S.~Osentoski, G.~Jay, and O.~C. Jenkins.
\newblock \href{https://doi.org/10.1109/ICHR.2010.5686284}{Learning from
  demonstration using a multi-valued function regressor for time-series data}.
\newblock In \emph{2010 10th IEEE-RAS International Conference on Humanoid
  Robots}, pages 328--333, Dec 2010.

\bibitem[Chiappa and Peters(2010)]{Chiappa10}
Silvia Chiappa and Jan~R. Peters.
\newblock
  \href{https://papers.nips.cc/paper/4109-movement-extraction-by-detecting-dynamics-switches-and-repetitions.pdf}{Movement
  extraction by detecting dynamics switches and repetitions}.
\newblock In J.~D. Lafferty, C.~K.~I. Williams, J.~Shawe-Taylor, R.~S. Zemel,
  and A.~Culotta, editors, \emph{Advances in Neural Information Processing
  Systems 23}, pages 388--396. Curran Associates, Inc., 2010.

\bibitem[Dantam et~al.(2012)Dantam, Essa, and Stilman]{Dantam12}
N.~Dantam, I.~Essa, and M.~Stilman.
\newblock \href{https://doi.org/10.1109/IROS.2012.6385749}{Linguistic transfer
  of human assembly tasks to robots}.
\newblock In \emph{2012 IEEE/RSJ International Conference on Intelligent Robots
  and Systems}, pages 237--242, Oct 2012.

\bibitem[De~la Higuera(2010)]{higuera2010grammatical}
Colin De~la Higuera.
\newblock \emph{Grammatical inference: learning automata and grammars}.
\newblock Cambridge University Press, 2010.

\bibitem[Dixon and Khosla(2004)]{Dixon04}
K.~R. Dixon and P.~K. Khosla.
\newblock \href{https://doi.org/10.1109/ROBOT.2004.1308881}{Trajectory
  representation using sequenced linear dynamical systems}.
\newblock In \emph{IEEE International Conference on Robotics and Automation,
  2004. Proceedings. ICRA '04. 2004}, volume~4, pages 3925--3930 Vol.4, April
  2004.

\bibitem[Gaunt et~al.(2017)Gaunt, Brockschmidt, Kushman, and Tarlow]{gaunt2017}
Alexander~L. Gaunt, Marc Brockschmidt, Nate Kushman, and Daniel Tarlow.
\newblock \href{https://arxiv.org/pdf/1611.02109.pdf}{Differentiable Programs
  with Neural Libraries}.
\newblock In \emph{Proceedings of the 34th International Conference on Machine
  Learning}, volume~70 of \emph{Proceedings of Machine Learning Research},
  pages 1213--1222, International Convention Centre, Sydney, Australia, August
  2017. PMLR.

\bibitem[Goebel et~al.(2009)Goebel, Sanfelice, and Teel]{goebel2009hybrid}
Rafal Goebel, Ricardo~G Sanfelice, and Andrew~R Teel.
\newblock \href{https://doi.org/10.1109/MCS.2008.931718}{Hybrid dynamical
  systems}.
\newblock \emph{IEEE Control Systems}, 29\penalty0 (2):\penalty0 28--93, 2009.

\bibitem[Grollman and Jenkins(2008)]{grollman2008sparse}
Daniel~H Grollman and Odest~Chadwicke Jenkins.
\newblock \href{https://doi.org/10.1109/ROBOT.2008.4543716}{Sparse incremental
  learning for interactive robot control policy estimation}.
\newblock In \emph{Robotics and Automation, 2008. ICRA 2008. IEEE International
  Conference on}, pages 3315--3320. IEEE, 2008.

\bibitem[Gulwani et~al.(2017)Gulwani, Polozov, Singh,
  et~al.]{gulwani2017program}
Sumit Gulwani, Oleksandr Polozov, Rishabh Singh, et~al.
\newblock
  \href{https://www.microsoft.com/en-us/research/wp-content/uploads/2017/10/program_synthesis_now.pdf}{Program
  synthesis}.
\newblock \emph{Foundations and Trends{\textregistered} in Programming
  Languages}, 4\penalty0 (1-2):\penalty0 1--119, 2017.

\bibitem[Harnad(1990)]{HARNAD1990335}
Stevan Harnad.
\newblock \href{https://doi.org/10.1016/0167-2789(90)90087-6}{The symbol
  grounding problem}.
\newblock \emph{Physica D: Nonlinear Phenomena}, 42\penalty0 (1):\penalty0 335
  -- 346, 1990.

\bibitem[Kaufmann et~al.(2018)Kaufmann, Loquercio, Ranftl, Dosovitskiy, Koltun,
  and Scaramuzza]{kaufmann18}
Elia Kaufmann, Antonio Loquercio, Rene Ranftl, Alexey Dosovitskiy, Vladlen
  Koltun, and Davide Scaramuzza.
\newblock \href{http://rpg.ifi.uzh.ch/docs/CORL18_Kaufmann.pdf}{Deep Drone
  Racing: Learning Agile Flight in Dynamic Environments}.
\newblock In \emph{Proceedings of the 2nd Annual Conference on Robot Learning},
  13--15 Nov 2018.

\bibitem[Kong et~al.(1994)Kong, Liu, and Wong]{kong1994sequential}
Augustine Kong, Jun~S Liu, and Wing~Hung Wong.
\newblock \href{https://www.jstor.org/stable/pdf/2291224.pdf}{Sequential
  imputations and Bayesian missing data problems}.
\newblock \emph{Journal of the American statistical association}, 89\penalty0
  (425):\penalty0 278--288, 1994.

\bibitem[Konidaris and Barto(2009)]{konidaris2009skill}
George Konidaris and Andrew~G Barto.
\newblock
  \href{http://papers.nips.cc/paper/3683-skill-discovery-in-continuous-reinforcement-learning-domains-using-skill-chaining}{Skill
  discovery in continuous reinforcement learning domains using skill chaining}.
\newblock In \emph{Advances in neural information processing systems}, pages
  1015--1023, 2009.

\bibitem[Lake et~al.(2015)Lake, Salakhutdinov, and Tenenbaum]{lake2015human}
Brenden~M Lake, Ruslan Salakhutdinov, and Joshua~B Tenenbaum.
\newblock \href{ttp://science.sciencemag.org/content/350/6266/1332}{Human-level
  concept learning through probabilistic program induction}.
\newblock \emph{Science}, 350\penalty0 (6266):\penalty0 1332--1338, 2015.

\bibitem[Lake et~al.(2017{\natexlab{a}})Lake, Ullman, Tenenbaum, and
  Gershman]{lake2017building}
Brenden~M Lake, Tomer~D Ullman, Joshua~B Tenenbaum, and Samuel~J Gershman.
\newblock \href{https://doi.org/10.1017/S0140525X16001837}{Building machines
  that learn and think like people}.
\newblock \emph{Behavioral and Brain Sciences}, 40, 2017{\natexlab{a}}.

\bibitem[Lake et~al.(2017{\natexlab{b}})Lake, Ullman, Tenenbaum, and
  Gershman]{lake_ullman_tenenbaum_gershman_2017}
Brenden~M. Lake, Tomer~D. Ullman, Joshua~B. Tenenbaum, and Samuel~J. Gershman.
\newblock \href{https://doi.org/10.1017/S0140525X17001224}{Ingredients of
  intelligence: From classic debates to an engineering roadmap}.
\newblock \emph{Behavioral and Brain Sciences}, 40:\penalty0 e281,
  2017{\natexlab{b}}.

\bibitem[L{\'a}zaro-Gredilla et~al.(2019)L{\'a}zaro-Gredilla, Lin, Guntupalli,
  and George]{Lazaro-Gredillaeaav3150}
Miguel L{\'a}zaro-Gredilla, Dianhuan Lin, J.~Swaroop Guntupalli, and Dileep
  George.
\newblock
  \href{http://robotics.sciencemag.org/content/4/26/eaav3150.full.pdf}{Beyond
  imitation: Zero-shot task transfer on robots by learning concepts as
  cognitive programs}.
\newblock \emph{Science Robotics}, 4\penalty0 (26), 2019.

\bibitem[Levine and Abbeel(2014)]{levine2014learning}
Sergey Levine and Pieter Abbeel.
\newblock
  \href{http://papers.nips.cc/paper/5444-learning-neural-network-policies-with-guided-policy-search-under-unknown-dynamics.pdf}{Learning
  neural network policies with guided policy search under unknown dynamics}.
\newblock In \emph{Advances in Neural Information Processing Systems}, pages
  1071--1079, 2014.

\bibitem[Levine et~al.(2016)Levine, Finn, Darrell, and Abbeel]{levine2016end}
Sergey Levine, Chelsea Finn, Trevor Darrell, and Pieter Abbeel.
\newblock
  \href{http://www.jmlr.org/papers/volume17/15-522/15-522.pdf}{End-to-end
  training of deep visuomotor policies}.
\newblock \emph{The Journal of Machine Learning Research}, 17\penalty0
  (1):\penalty0 1334--1373, 2016.

\bibitem[Lieberman et~al.(2006)Lieberman, Patern{\`o}, Klann, and
  Wulf]{lieberman2006enduser}
Henry Lieberman, Fabio Patern{\`o}, Markus Klann, and Volker Wulf.
\newblock \emph{End-User Development: An Emerging Paradigm}.
\newblock Springer Netherlands, 2006.

\bibitem[Marcus(2003)]{marcus2001algebraic}
Gary Marcus.
\newblock \emph{The algebraic mind: Integrating connectionism and cognitive
  science}.
\newblock MIT Press, 2003.

\bibitem[Menghi et~al.(2019)Menghi, Tsigkanos, Pelliccione, Ghezzi, and
  Berger]{menghi2019specification}
Claudio Menghi, Christos Tsigkanos, Patrizio Pelliccione, Carlo Ghezzi, and
  Thorsten Berger.
\newblock \href{https://arxiv.org/pdf/1901.02077.pdf}{Specification Patterns
  for Robotic Missions}.
\newblock \emph{arXiv preprint arXiv:1901.02077}, 2019.

\bibitem[Niekum and Barto(2011)]{Niekum11}
Scott Niekum and Andrew~G. Barto.
\newblock
  \href{https://papers.nips.cc/paper/4238-clustering-via-dirichlet-process-mixture-models-for-portable-skill-discovery.pdf}{Clustering
  via Dirichlet Process Mixture Models for Portable Skill Discovery}.
\newblock In J.~Shawe-Taylor, R.~S. Zemel, P.~L. Bartlett, F.~Pereira, and
  K.~Q. Weinberger, editors, \emph{Advances in Neural Information Processing
  Systems 24}, pages 1818--1826. Curran Associates, Inc., 2011.

\bibitem[Niekum et~al.(2012)Niekum, Osentoski, Konidaris, and Barto]{Niekum12}
Scott Niekum, S.~Osentoski, George Konidaris, and Andrew~G. Barto.
\newblock \href{https://doi.org/10.1109/IROS.2012.6386006}{Learning and
  generalization of complex tasks from unstructured demonstrations}.
\newblock In \emph{2012 IEEE/RSJ International Conference on Intelligent Robots
  and Systems}, pages 5239--5246, Oct 2012.

\bibitem[Niekum et~al.(2013)Niekum, Chitta, Barto, Marthi, and
  Osentoski]{Niekum-RSS-13}
Scott Niekum, Sachin Chitta, Andrew Barto, Bhaskara Marthi, and Sarah
  Osentoski.
\newblock \href{http://www.roboticsproceedings.org/rss09/p48.pdf}{Incremental
  Semantically Grounded Learning from Demonstration}.
\newblock In \emph{Proceedings of Robotics: Science and Systems}, Berlin,
  Germany, June 2013.

\bibitem[Niekum et~al.(2015)Niekum, Osentoski, Konidaris, Chitta, Marthi, and
  Barto]{niekum2015learning}
Scott Niekum, Sarah Osentoski, George Konidaris, Sachin Chitta, Bhaskara
  Marthi, and Andrew~G Barto.
\newblock
  \href{https://journals.sagepub.com/doi/abs/10.1177/0278364914554471}{Learning
  grounded finite-state representations from unstructured demonstrations}.
\newblock \emph{The International Journal of Robotics Research}, 34\penalty0
  (2):\penalty0 131--157, 2015.

\bibitem[Pastor et~al.(2009)Pastor, Hoffmann, Asfour, and
  Schaal]{pastor2009learning}
Peter Pastor, Heiko Hoffmann, Tamim Asfour, and Stefan Schaal.
\newblock \href{https://doi.org/10.1109/ROBOT.2009.5152385}{Learning and
  generalization of motor skills by learning from demonstration}.
\newblock In \emph{Robotics and Automation, 2009. ICRA'09. IEEE International
  Conference on}, pages 763--768. IEEE, 2009.

\bibitem[Penkov and Ramamoorthy(2017)]{penkov2017explaining}
Svetlin Penkov and Subramanian Ramamoorthy.
\newblock \href{https://arxiv.org/pdf/1708.00376.pdf}{Explaining Transition
  Systems through Program Induction}.
\newblock \emph{arXiv preprint arXiv:1705.08320}, 2017.

\bibitem[Penkov and Ramamoorthy(2019)]{penkov2019learning}
Svetlin Penkov and Subramanian Ramamoorthy.
\newblock \href{https://openreview.net/forum?id=SJggZnRcFQ}{Learning
  Programmatically Structured Representations with Perceptor Gradients}.
\newblock In \emph{International Conference on Learning Representations}, 2019.

\bibitem[Penkov et~al.(2017)Penkov, Bordallo, and Ramamoorthy]{Penkov17}
Svetlin Penkov, Alejandro Bordallo, and Subramanian Ramamoorthy.
\newblock \href{https://doi.org/10.1109/ICRA.2017.7989697}{Physical symbol
  grounding and instance learning through demonstration and eye tracking}.
\newblock In \emph{2017 IEEE International Conference on Robotics and
  Automation (ICRA)}, pages 5921--5928, May 2017.

\bibitem[Pinto and Gupta(2016)]{pinto2016supersizing}
Lerrel Pinto and Abhinav Gupta.
\newblock
  \href{https://ieeexplore.ieee.org/stamp/stamp.jsp?arnumber=7487517}{Supersizing
  self-supervision: Learning to grasp from 50k tries and 700 robot hours}.
\newblock In \emph{Robotics and Automation (ICRA), 2016 IEEE International
  Conference on}, pages 3406--3413. IEEE, 2016.

\bibitem[Rajeswaran et~al.(2018)Rajeswaran, Kumar, Gupta, Vezzani, Schulman,
  Todorov, and Levine]{Rajeswaran-RSS-18}
Aravind Rajeswaran, Vikash Kumar, Abhishek Gupta, Giulia Vezzani, John
  Schulman, Emanuel Todorov, and Sergey Levine.
\newblock \href{http://roboticsproceedings.org/rss14/p49.html}{Learning Complex
  Dexterous Manipulation with Deep Reinforcement Learning and Demonstrations}.
\newblock In \emph{Proceedings of Robotics: Science and Systems}, Pittsburgh,
  Pennsylvania, June 2018.

\bibitem[Ranchod et~al.(2015)Ranchod, Rosman, and
  Konidaris]{ranchod2015nonparametric}
Pravesh Ranchod, Benjamin Rosman, and George Konidaris.
\newblock \href{https://doi.org/10.1109/IROS.2015.7353414}{Nonparametric
  bayesian reward segmentation for skill discovery using inverse reinforcement
  learning}.
\newblock In \emph{Intelligent Robots and Systems (IROS), 2015 IEEE/RSJ
  International Conference on}, pages 471--477. IEEE, 2015.

\bibitem[Selvaraju et~al.(2017)Selvaraju, Cogswell, Das, Vedantam, Parikh, and
  Batra]{selvaraju2017grad}
Ramprasaath~R Selvaraju, Michael Cogswell, Abhishek Das, Ramakrishna Vedantam,
  Devi Parikh, and Dhruv Batra.
\newblock \href{https://arxiv.org/abs/1610.02391}{Grad-cam: Visual explanations
  from deep networks via gradient-based localization}.
\newblock In \emph{2017 IEEE International Conference on Computer Vision
  (ICCV)}, pages 618--626. IEEE, 2017.

\bibitem[Sutton et~al.(1999)Sutton, Precup, and Singh]{sutton1999between}
Richard~S Sutton, Doina Precup, and Satinder Singh.
\newblock \href{https://doi.org/10.1016/S0004-3702(99)00052-1}{Between MDPs and
  semi-MDPs: A framework for temporal abstraction in reinforcement learning}.
\newblock \emph{Artificial intelligence}, 112\penalty0 (1-2):\penalty0
  181--211, 1999.

\bibitem[Tedrake(2009)]{Tedrake-RSS-09}
R.~Tedrake.
\newblock \href{http://roboticsproceedings.org/rss05/p3.html}{{LQR}-trees:
  Feedback motion planning on sparse randomized trees}.
\newblock In \emph{Proceedings of Robotics: Science and Systems}, Seattle, USA,
  June 2009.

\bibitem[Verma et~al.(2018)Verma, Murali, Singh, Kohli, and
  Chaudhuri]{verma2018programmatically}
Abhinav Verma, Vijayaraghavan Murali, Rishabh Singh, Pushmeet Kohli, and Swarat
  Chaudhuri.
\newblock \href{https://arxiv.org/pdf/1804.02477.pdf}{Programmatically
  Interpretable Reinforcement Learning}.
\newblock \emph{arXiv preprint arXiv:1804.02477}, 2018.

\bibitem[Yu et~al.(2018)Yu, Finn, Dasari, Xie, Zhang, Abbeel, and
  Levine]{Yu-RSS-18}
Tianhe Yu, Chelsea Finn, Sudeep Dasari, Annie Xie, Tianhao Zhang, Pieter
  Abbeel, and Sergey Levine.
\newblock \href{http://roboticsproceedings.org/rss14/p02.html}{One-Shot
  Imitation from Observing Humans via Domain-Adaptive Meta-Learning}.
\newblock In \emph{Proceedings of Robotics: Science and Systems}, Pittsburgh,
  Pennsylvania, June 2018.

\bibitem[Zeiler and Fergus(2014)]{zeiler2014visualizing}
Matthew~D Zeiler and Rob Fergus.
\newblock \href{https://arxiv.org/abs/1311.2901}{Visualizing and understanding
  convolutional networks}.
\newblock In \emph{European conference on computer vision}, pages 818--833.
  Springer, 2014.

\bibitem[Zhu et~al.(2018)Zhu, Wang, Merel, Rusu, Erez, Cabi, Tunyasuvunakool,
  Kramár, Hadsell, de~Freitas, and Heess]{Zhu-RSS-18}
Yuke Zhu, Ziyu Wang, Josh Merel, Andrei Rusu, Tom Erez, Serkan Cabi, Saran
  Tunyasuvunakool, János Kramár, Raia Hadsell, Nando de~Freitas, and Nicolas
  Heess.
\newblock \href{http://roboticsproceedings.org/rss14/p12.pdf}{Reinforcement and
  Imitation Learning for Diverse Visuomotor Skills}.
\newblock In \emph{Proceedings of Robotics: Science and Systems}, Pittsburgh,
  Pennsylvania, June 2018.

\end{thebibliography}

\end{document}